\documentclass{sig-alternate-2013}

\newfont{\mycrnotice}{ptmr8t at 7pt}
\newfont{\myconfname}{ptmri8t at 7pt}
\let\crnotice\mycrnotice%
\let\confname\myconfname%

\permission{Permission to make digital or hard copies of all or part of this work for personal or classroom use is granted without fee provided that copies are not made or distributed for profit or commercial advantage and that copies bear this notice and the full citation on the first page. Copyrights for components of this work owned by others than the author(s) must be honored. Abstracting with credit is permitted. To copy otherwise, or republish, to post on servers or to redistribute to lists, requires prior specific permission and/or a fee. Request permissions from Permissions@acm.org.}
\conferenceinfo{SIGIR '15,}{August 9--13, 2015, Santiago, Chile. \\
{\mycrnotice{Copyright is held by the owner/author(s). Publication rights licensed to ACM.}}}
\CopyrightYear{2015}
\crdata{978-1-4503-3621-5/15/08...\$15.00\\DOI: http://dx.doi.org/10.1145/2766462.2767789} 
\usepackage{cite}
\usepackage{graphicx}
\usepackage{subfig}
\usepackage{url}
\usepackage{xspace}
\usepackage{algorithmic}
\usepackage{algorithm}
\usepackage{multirow}

\newdef{defn}{Definition}

\usepackage{eurosym}
\usepackage{amstext} 
\DeclareRobustCommand{\officialeuro}{%
  \ifmmode\expandafter\text\fi
  {\fontencoding{U}\fontfamily{eurosym}\selectfont e}}

\def\ie{\emph{i.e.,~}}
\def\eg{\emph{e.g.,~}}

\newcommand{\appname}{{\sc MQSearch}\xspace}

\newcounter{Lcount}
\newcommand{\numsquishlist}{
  \begin{list}{\arabic{Lcount}. }
   { \usecounter{Lcount}
 \setlength{\itemsep}{0pt}      \setlength{\parsep}{3pt}
     \setlength{\topsep}{3pt}       \setlength{\partopsep}{0pt}
     \setlength{\leftmargin}{2.7em} \setlength{\labelwidth}{1em}
     \setlength{\labelsep}{0.5em} } }
\newcommand{\numsquishend}{\end{list}}

\newcommand{\squishlist}{
  \begin{list}{$\bullet$}
   { \setlength{\itemsep}{0pt}      \setlength{\parsep}{3pt}
     \setlength{\topsep}{3pt}       \setlength{\partopsep}{0pt}
     \setlength{\leftmargin}{1.5em} \setlength{\labelwidth}{1em}
     \setlength{\labelsep}{0.5em} } }
\newcommand{\squishend}{\end{list}}

\usepackage{color, colortbl}
\definecolor{Gray}{gray}{0.9}
\definecolor{LightCyan}{rgb}{0.88,1,1}

\begin{document}

\title{Mining Measured Information from Text}

\numberofauthors{1}


\author{
\alignauthor
Arun S. Maiya, Dale Visser, and Andrew Wan\\
       \affaddr{Institute for Defense Analyses --- Alexandria, VA, USA}\\
       \email{\{amaiya, dvisser, awan\}@ida.org}\\
}

\newcounter{copyrightbox}

\maketitle
\begin{abstract}

We present an approach to extract {\em measured information} from text (\eg a $1370~^{\circ}C$ melting point, a BMI greater than 29.9 kg/m$^2$).  Such extractions are critically important across a wide range of domains --- especially those involving search and exploration of scientific and technical documents.   We first propose a rule-based entity extractor to mine {\em measured quantities} (\ie a numeric value paired with a measurement unit), which supports a vast and comprehensive set of both common and obscure measurement units.  Our method is highly robust and can correctly recover valid measured quantities even when significant errors are introduced through the process of converting document formats like PDF to plain text. Next, we describe an approach to extracting the {\em properties} being measured (\eg the property {\em pixel pitch} in the phrase ``a pixel pitch as high as $352~\mu m$'').  Finally, we present \appname:  the realization of a search engine with full support for {\em measured information}. 
\end{abstract}


\section{Introduction and Motivation}
\label{sec:intro}

Scientific and technical documents describe methods and results using {\em measured quantities}: a numeric value paired with a unit of measurement.  Examples of text snippets containing such measured quantities include: 

\squishlist
\item  {\em average gravity curvature} $\zeta = (1.3999 \pm 0.003) \times 10^{-5} s^{-2}m^{-1}$
\item {\em $12~^{\circ}C$ melting point}
\item {\em distance from Earth to the Sun is $9.3 \times 10^7$ miles}
\item {\em average responsivity as low as $6.2$~pA/K}
\squishend

Note that these measured quantities (\eg $6.2$~pA/K) are typically associated with a specific {\em measured property} (\eg average responsivity).  In this paper, we study ways in which to extract these kinds of {\em measured information} from documents.\footnote{We define {\em measured information} as {\em measured quantities} and the {\em measured properties} to which they are associated.} The mining of such information is critically important across many domains --- especially those involving search and exploration of scientific and technical articles.  For instance, an optics researcher may wish to know if the performance of Nd:YAG laser-pumped KTP parametric oscillators has ever been tested at wavelengths longer than $2.4~\mu m$.  Full-text search engines using inverted indexes allow ad hoc queries on terms such as ``KTP parametric oscillator'', but the ability to further filter search results based on wavelengths greater than $2.4~\mu m$ is {\em not} typically supported.  To accomplish this, one must first identify and extract valid {\em measured quantities} (\eg $2.4~\mu m$) in unstructured text and, then, identify and extract the {\em properties} being measured (\eg {\em wavelength}).  These extractions could then be stored in the index of a search engine in a way that supports subsequent document queries on measured information (\eg faceted navigation, numeric range queries).  

Surprisingly, there is very little existing work on how best to realize this process. Lines of research most closely related to the present work include extracting numerical attributes (\eg \cite{Bakalov2011Scad,Davidov2010Extraction}), supporting numerical document queries (\eg \cite{Seidl2003Numerical,Fontoura2006Inverted}), and formula identification (\eg \cite{Lin2011Mathematical}). However, none of these existing works address the comprehensive extraction of and search for measured information in document data, as described above.  Indeed, numerous challenges exist in such scenarios. Many widely-used, full-text search engines (\eg Apache Solr) convert the original document format to {\em plain text} prior to indexing and storage --- an {\em extremely} error-ridden process.  For instance, in the extracted text, exponents are typically lost (\eg $10^5$ becomes $105$, $s^{-2}$ becomes $s$--$2$).  Moreover, the conversion of some characters can be highly inconsistent and unpredictable.  A simple minus sign can be converted to a range of different dash characters or even ``garbage'' characters.  The same is true for other symbols such as $\mu$, multiplication signs, and degree symbols.  It is this inconsistent and error-ridden text, then, that is ultimately stored in the index of the search engine making it virtually impossible to adequately locate documents by measured quantities. Without the correct identification of measured quantities, it is virtually impossible to identify {\em properties} being measured, which are critical in efficiently navigating scientific and technical articles for state-of-the-art information. In general, there is a great deal of heterogeneity in how {\em measured quantities} and {\em measured properties} appear in text -- both naturally and through corruption.  This, then, motivates the current investigation of how best to extract such information.

Recent studies \cite{Chiticariu2013Rulebased,Gupta2014SPIED} have revealed that rule-based approaches to information extraction tend to be more effective, interpretable, and customizable than state-of-the-art machine learning approaches. We employ rule-based extraction methods in this work.  Our contributions are as follows:

\squishlist
\item We propose and describe a rule-based entity extractor to identify  {\em measured quantities} in unstructured text documents.  Our method includes an error-correcting procedure that recovers from aforementioned text conversion errors by 1) {\em reverse engineering} the corrupted and mangled measured quantities back to their original, correct form and 2) {\em standardizing} this form for storage in an inverted index and subsequent query processing.
\item Using these extracted measured quantities, we show how to further extract the {\em measured properties} to which they are associated.  
\item Finally, we present \appname:  the realization of a search engine with full support for {\em measured information}.  \appname is a facet-based navigation system that allows users to navigate large document sets based on measured quantities, measured properties, and the topics and themes to which they are associated.  To the best of our knowledge, no other search engine in existence fully supports such a capability.
\squishend
\noindent
We begin with describing the extraction of {\em measured quantities}.
  
\begin{table*}[thb]
\centering
{\scriptsize
\begin{tabular}{|l|l|l|} \hline
{\bf Rule}         &  {\bf Pattern}   & {\bf Example Matches}            \\ \hline  \hline

{\em 1) number}      &  [$+-$]?(\textbackslash d((\textbackslash d?\textbackslash d?[, ]\textbackslash d\{2,3\}([, ]\textbackslash d\{2,3\})*)|\textbackslash d*))(\textbackslash .(\textbackslash d[\textbackslash d\textbackslash s]*\textbackslash d|\textbackslash d))? & 1000.05, +5, -0.2, and 1,000   \\ \hline
{\em 2) number {\scriptsize (leading point)}}    &  [+-]?\textbackslash .\textbackslash d(\textbackslash d\textbackslash d(\textbackslash s\textbackslash d\{3\})+(\textbackslash s\textbackslash d\{1,3\})?|\textbackslash d*)   &  -.98, .04, +.755        \\ \hline
{\em 3) error }      & (\textbackslash s\{0,2\} $\pm$ \textbackslash s\{0,2\}[\textbackslash d.]+)? & $\pm$0.003 in  ``$1.3999 \pm 0.003$''   \\ \hline
  {\em 4) sci. notation.}           &  (\textbackslash s*[eE]|\textbackslash s*([xX$\times$ ])\textbackslash s*10 *\textbackslash \textasciicircum? [+-]?\textbackslash d+)? & \eg forms of $\times 10^{-5}$:  $\times 105$, e-5, E-5  \\ \hline
{\em 5) unit}    & \eg [fpn$\mu$mcdk]?m([\textbackslash\textasciicircum]?[2-6] | [\textbackslash-][1-6])   --- {\bf $m^{\#}$ normalized to $m$\^{}\#}  & $\mu m$, $m$--1 ($m^{-1}$), $cm2$ ($cm^{2}$), $cm$\textasciicircum 2                    \\ \hline
   {\em 6) connector}      & (\textbackslash s?/\textbackslash s? | [Pp]er |-per-| [-\textbackslash s$\times$ $\cdot$*])? & per, /, $\cdot$, $\times$    \\ \hline
   {\em 7) compound unit}      & <unit>(<connector><unit>)+ & km/h, kilometer per hour,km$\cdot h^{-1}$    \\

     \hline
\end{tabular}
\caption{{\footnotesize {\bf [MQE Rules.]}  Simplified forms of some rules for extraction of {\em measured quantities}.}}
\label{tab:mqe}
}
\vskip 0.1in
\end{table*}

~\\

\section{Measured Quantities}
\label{sec:mq}
We view {\em measured quantities} as a 5-tuple of the form: ({\em sign}, \underline{{\em number}}, {\em error}, {\em scientific notation}, \underline{{\em units}}), where underlined elements are mandatory and others are optional.  As an example, a team of researchers in Italy recently reported the first direct measurement of gravity's curvature as $ (1.3999 \pm 0.003) \times 10^{-5} s^{-2}m^{-1}$ \cite{Rosi2015Measurement}. The corresponding 5-tuple representation of this\footnote{Since there is no explicit sign in this example, the first element is left empty.} is:
\begin{verse}
{\footnotesize (<empty>, $1.3999$, $0.003$, $10^{-5}$,  $s^{-2}m^{-1}$)}.
\end{verse}
5-tuples such as this are populated using a series of extraction rules that operate on individual sentences. These rules fall into four broad categories: 1) pre-processing, 2) units, 3) quantities, and 4) post-processing.  Simplified forms of some of the rules for units and quantities are shown in Table \ref{tab:mqe}.\footnote{Rules are shown in Perl-like syntax, the de facto standard for regular expressions.}  We refer to the algorithm implementing such rules as {\em Measured Quantity Extractor} or {\bf MQE}.  We begin with pre-processing rules.

~\\
\noindent
{\bf Pre-Processing.}  As mentioned previously, when extracting text from various document formats (\eg PDF, MS Office), characters often appear inconsistently.  Minus signs, multiplication signs (\eg $\times$, $\cdot$), equal-like symbols (\eg $\approx$, $\simeq$, $\cong$), degree symbols, and the $\mu$ character can appear in a variety of ways or, in some cases, as ``garbage'' characters. For instance, minus can appear as the {\em en dash} character or appear corrupted as \symbol{226}$\euro$. Pre-processing rules identify these variations in text and perform the necessary normalization for accurate extraction of units and quantities.

~\\
\noindent
{\bf Units.}    A {\em measurement unit} preceded by a numeric string conforming to the 5-tuple structure is the base  indicator of a {\em measured quantity}. Thus, to identify valid {\em measured quantities}, we require a comprehensive ontology of units.  We obtained an initial units ontology from the OBO Foundry,\footnote{\url{http://www.obofoundry.org/}} but this was quite incomplete.  We, then, expanded the ontology using largely public sources (\eg \url{convert-me.com}, DoD technical reports, Physical Review, Nature Communications). Each unit has an associated rule.  An example rule for {\em m} (\ie symbol for meters) is shown in Rule 5 of Table \ref{tab:mqe}.  Note that such rules include optional prefixes for submultiples and multiples (\eg $\mu$ before {\em m}, {\em kilo} before {\em meter}).  Unit rules, when combined with pre-processing rules described previously, can accurately extract units under a range of noisy conditions.    For instance, the corrupted unit $m$\symbol{226}$\euro 1$ is correctly recovered as $m^{-1}$ by MQE.  Finally, as shown in Rules 6 and 7, compound units are also supported (\eg km/h, kilometer per hour, $s^{-2}\cdot m^{-1}$).

~\\
\noindent
{\bf Quantities.}  Like units, quantities (\ie numbers with optional error ranges and scientific notation) can appear in a range of ways due to both corruptions and natural variation.  These variations are collectively captured by rules such as those shown in Table \ref{tab:mqe} (\ie Rules 1-4), which populate the remainder of the 5-tuple structure.  As shown in Table \ref{tab:mqe}, such rules capture a wide range of quantity formats (\eg $10,000$ with a comma, $ 1.3999 \pm 0.003 \times 10^{-5}$ with both an error range and scientific notation, $1.23 \times 105$ with lost exponent in $10^5$). To support numeric range queries, extracted quantities are standardized prior to storage in a search engine index (\eg the extracted quantity $ 1.3999 \pm 0.003 \times 10^{-5}$ is stored simply as $0.000013999$) \cite{Schindler2008Generic}.

~\\
\noindent
{\bf Post-Processing.}
We have already seen that text extracted from various document formats can be noisy.  For instance, information from tables, headers, and figures can sometimes result in seemingly random sequences of numbers and letters in extracted text.  In some cases, such information can erroneously be picked up by aforementioned rules as {\em measured quantities}.  This is especially true for single letter units (\eg {\em m} for meters, {\em A} for Ampere, etc.).  Post-processing rules are employed to reject such extractions and minimize false positives.  Examples of such rejection rules include context-based rules (\eg reject when preceded by ``Table'' or ``Figure''), repetition-based rules such as rejecting compound units consisting of repeated single letter units (\eg 3 AJmm), and allowing a dash only between certain quantities and units (\eg {\em 10-cm} is okay but not {\em 10-A}).

~\\
As we will show in Section \ref{sec:eval}, when used in combination, these rules collectively enable highly accurate extractions of {\em measured quantities} -- which, in turn, can be exploited to extract the {\em properties} being measured, as described next.

\begin{table*}[htb]
\centering
{\scriptsize
\begin{tabular}{|l|l|} \hline
{\bf Pattern} &  {\bf Example Matches} ~~~~~~(two examples shown for each rule)\\ \hline \hline
\multirow{1}{*}{\textcolor{red}{NP} SYM\{0,2\} EQ \textcolor{blue}{mq}} &  1) {\scriptsize\em \textcolor{red}{gravity curvature} \underline{$\zeta =$} \textcolor{blue}{$ 1.4\times10^{-5} s^{-2}m^{-1}$}}~~2) {\scriptsize\em \textcolor{red}{floor area} \underline{$\approx$}  \textcolor{blue}{$32 m^2$}}\\ \hline
\multirow{1}{*}{\textcolor{blue}{mq} IN? \textcolor{red}{NP}} &  1) {\scriptsize\em a \textcolor{blue}{$352~\mu m$} \textcolor{red}{pixel pitch}}~~~~~~~~~~~~~~~~~~~~~~~~~~~~ 2) {\scriptsize\em \textcolor{blue}{$50 mL$} \underline{of} \textcolor{red}{30\% fuming sulfuric acid}}\\ \hline
\textcolor{red}{NP IN DT? NP} VP+ (TO|IN|RB|JJ)* \textcolor{blue}{mq} & 1) {\em \textcolor{red}{strength of panel} \underline{was set to} \textcolor{blue}{9 ksi}}~~~~~~~~~~~~2) {\em \textcolor{red}{freq. of scans} \underline{was roughly} \textcolor{blue}{300 Hz}}\\ \hline
\multirow{1}{*}{\textcolor{red}{NP} (IN DT? NP)* VP+ (IN|TO|RB|JJ)* \textcolor{blue}{mq}} & 1) {\em  \textcolor{red}{pixel pitch} \underline{employed was} \textcolor{blue}{$352~\mu m$}.}~~~~~~~~~~~2) {\em \textcolor{red}{panel strength} \underline{was recorded at} \textcolor{blue}{$9~ksi$}.}\\ \hline
\multirow{1}{*}{\textcolor{red}{NP} (CC|IN|TO|RB|JJ)* \textbackslash(?\textcolor{blue}{mq}\textbackslash)?} & 1) {\em \textcolor{red}{wavelengths} \underline{of at least} \textcolor{blue}{$2.4~\mu m$}}~~~~~~~~~~~~~~~~2) {\em \textcolor{red}{panel strength} (\textcolor{blue}{$9~ksi$})} \\ \hline
\end{tabular}
\caption{{\bf [MPE Rules.]} {\footnotesize Simplified forms of some syntactic patterns to extract {\em measured properties}.}}
\label{tab:pos}
}
\vskip -0.1in
\end{table*}

\section{Measured Properties}
\label{sec:mp}

We now turn our attention to the extraction of {\em measured properties}.  To better illustrate the problem, we show several example snippets containing measured quantities.  In each example, the {\em measured quantity} is shown in \textcolor{blue}{blue}, the {\em property} being measured is highlighted in \textcolor{red}{red}, and the characters connecting them are \underline{underlined}:
\squishlist
\item {\footnotesize\em a \textcolor{red}{pixel pitch} \underline{as high as roughly} \textcolor{blue}{$352~\mu m$}}
\item {\footnotesize\em a \textcolor{blue}{$352~\mu m$} \textcolor{red}{pixel pitch}}
\item {\footnotesize\em The \textcolor{red}{pixel pitch} \underline{employed was} \textcolor{blue}{$352~\mu m$}.}
\item  {\footnotesize\em \textcolor{red}{average gravity curvature}} \underline{$\zeta =$}\textcolor{blue}{$ (1.3999 \pm 0.003) \times 10^{-5} s^{-2}m^{-1}$}
\item {\footnotesize\em with \textcolor{blue}{$50 mL$} \underline{of} \textcolor{red}{30\% fuming sulfuric acid}}
\item  {\footnotesize\em \textcolor{red}{size} \underline{$\cong$}  \textcolor{blue}{$0.1 m^2$}}
\item {\footnotesize\em \textcolor{red}{frequency of longitudinal scan} \underline{was approximately} \textcolor{blue}{300 Hz}.}
\item {\footnotesize\em a \textcolor{red}{nominal current density} \underline{of} \textcolor{blue}{1.3 A/cm$^{2}$} \underline{to} \textcolor{blue}{0.03 A/cm$^{2}$}}
\item {\footnotesize\em \textcolor{red}{panel strength} \underline{lower than} \textcolor{blue}{8.90 ksi} (\textcolor{blue}{61.4 MPa})}
\item {\footnotesize\em \textcolor{red}{wavelengths} \underline{at least} \textcolor{blue}{$2.4~\mu m$}}
\item {\footnotesize\em \textcolor{red}{large fields} \underline{of about, or above} \textcolor{blue}{10 kV/cm}}
\squishend

From just the examples shown, it is easy to see that there is an extremely high degree of variability in the words connecting a {\em measured property} with a {\em measured quantity}. These examples represent just a small sample of the many possible variations. However, upon closer inspection, we find that this variability can be reduced to a small number of syntactic patterns based on part-of-speech (POS) that capture most scenarios.  Table \ref{tab:pos} shows some syntactic patterns that we employ to extract {\em measured properties}.  We refer to the extractor applying such syntactic rules as {\em Measured Property Extractor} or {\bf MPE}.

In Table \ref{tab:pos}, noun phrases shown in red (\ie \textcolor{red}{NP}) are extracted and taken as the {\em measured property}. Measured quantities are represented in blue by \textcolor{blue}{mq}. The EQ tag represents all symbols related to '=' (\eg $\approx$, $\simeq$). The SYM tag matches one or two character symbols (\eg a greek letter).  Other symbols (\eg JJ, RB, IN, CC, VP) are part-of-speech tags in Penn Treebank format.  Note that tags such as RB (\ie an adverb) should be taken to include variations such as the comparative and superlative forms.  This is not explicitly shown for reasons of brevity.  This small set of patterns matches a very wide range of possible phrase combinations for {\em measured properties} and are executed sequentially in the order shown.  We implemented MPE using the Brill part-of-speech tagger \cite{Brill1992Simple}. As we will show in the next section, the accuracy with which our algorithms are able to extract measured properties and measured quantities is remarkable --- especially given the aforementioned issues with noisy and corrupted input text.

\section{Experimental Evaluation}
\label{sec:eval}
Since our research is sponsored by the U.S. Department of Defense (DoD), we evaluate our approach on a text corpus consisting of 40,807 unclassified research reports published in the 2008-2010 time frame and hosted by the Defense Technical Information Center (DTIC).   This rich collection describes a wide range of research funded by the DoD spanning numerous fields from engineering and physical science to biomedical research and social science.   The DTIC documents considered in this paper have been approved for public release and unlimited distribution.  All documents are in PDF format, and text was extracted from them using the \mbox{\texttt{pdftotext}} utility.\footnote{\url{http://www.foolabs.com/xpdf/home.html}} From this collection, we generated samples using the following procedure. To evaluate the ability of MQE to extract measured quantities, we sampled uniform random sentences from the population of all sentences containing a numeric value. By examining sentences with a number (but not necessarily a measurement unit), we are able to accurately identify false negatives in addition to false positives. Next, to evaluate the ability of MPE to extract measured properties, we generated a random sample of sentences from the population of all sentences containing a measured quantity, as identified by MQE. We employed sample sizes of $1000$ and $500$ for MQE and MPE, respectively.  This produced sufficient 95\% confidence bounds on our estimates for precision and recall over the entire corpus.  Different fields employ different measures in different ways. By considering sentences sampled randomly in this fashion, we are able to evaluate our methods on text data that capture the diverse ways in which measured information is reported across different fields.  To the best of our knowledge, no other approaches exist for extracting such {\em measured information} from scientific and technical documents.  Thus, there are no appropriate baselines against which our methods can be compared. Table \ref{tab:results} shows the precision and recall estimates for both the measured quantity extractor and the measured property extractor over the entire corpus.

\begin{table}[thb]
\centering
{\footnotesize
\begin{tabular}{l|c|c} \hline \hline
{\bf Extractor}         &  {\bf Precision}    & {\bf Recall}          \\ \hline
 MQE    &   ($0.93$, $0.99$)             &  ($0.92$, $0.99$)                         \\
 MPE    &   ($0.93$, $0.97$)             & ($0.88$,  $0.94$)              \\
  \hline \hline
\end{tabular}
\caption{{\bf 95\% Confidence Intervals} for precision and recall when extracting {\em measured quantities} (using MQE) and {\em measured properties} (using MPE) from the DTIC corpus.}
\label{tab:results}
}
\end{table}

As can be seen in the table, both MQE and MPE perform extraordinarily well in extracting {\em measured quantities} and the {\em properties} they describe from documents across disparate fields.  Having demonstrated the success with which {\em measured information} can be mined, we now demonstrate how these extractions can be exploited in novel search applications.

\section{An Application:  MQSearch}
\begin{figure}[htb]
\begin{center}
\centerline{\fbox{\includegraphics[scale=0.35]{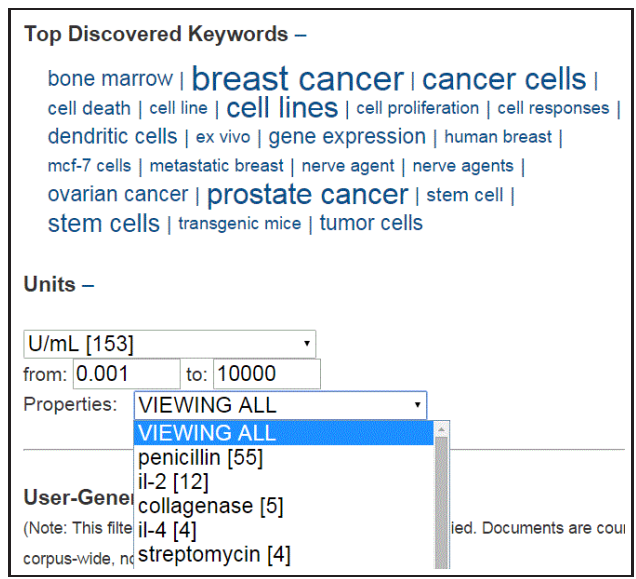}}}
\caption{{\footnotesize {\bf [\appname.]} The measurement unit {\em U/mL} is selected, which reveals the associated topics (\eg breast/prostate cancer), associated measured properties (\eg concentrations of {\em penicillin}), and associated quantity ranges (\ie {\em 0.001} to {\em 10,000}).}}
\label{fig:mqsearch}
\end{center}
\vskip -0.2in
\end{figure}

Here, we present \appname:  a realization of a search engine with full support for {\em measured information}.  \appname is implemented using Apache Solr\footnote{\url{http://lucene.apache.org/solr/}} and AJAX Solr\footnote{\url{https://github.com/evolvingweb/ajax-solr}}, both of which support full-text search, faceted navigation, and numeric range queries. During the process of indexing and ingesting the DTIC document set into our search engine, we apply our extractors to encountered text and store both {\em measured quantities} and {\em measured properties} in the search engine index.  In addition, the search engine performs keyphrase extraction on documents using the KERA algorithm described in \cite{Maiya2013Exploratory}. Using Solr filter queries, extracted keyphrases can be used to produce a tag cloud for any subset of the document set.  Figure \ref{fig:mqsearch} shows the faceted navigation panel of \appname, which allows users to filter documents based on discovered measurement units, quantity ranges, and measured properties.   In Figure \ref{fig:mqsearch}, the measurement unit {\em U/mL} is selected. We see that there are 153 documents (out of roughly 40,000) mentioning this unit with quantities ranging from $0.001$ U/mL to $10,000$ U/mL. The property most frequently measured in {\em U/mL} is {\em penicillin}.  From the tag cloud, we see that documents containing quantities measured in {\em U/mL} tend to cover topics such as breast cancer and prostate cancer research.\footnote{This cancer research was funded by U.S. Army MRMC through a congressionally directed research program.} The search results can be filtered further along any of these dimensions.  Filtering by LDA-discovered topics is also supported but not shown in the figure \cite{McCallum2002MALLET}.  To the best of our knowledge, ours is the first search engine with such support for {\em measured information}.

\label{sec:mp}

\section{Conclusion}

In this paper, we have proposed a demonstrably effective approach to extracting {\em measured information} from unstructured text data.  We showed both how to extract {\em measured quantities} and the {\em properties} being measured. We further demonstrated how such extractions might be used in a search engine for documents rich in measured information.  To the best of our knowledge, no other search engine in existence supports such functionality.  Our extraction methods have the potential to substantially improve search, navigation, and exploratory analysis of large or even massive collections of scientific and technical articles.  For future work, we plan on marrying our proposed approaches with other well-studied techniques for exploratory search.

{\scriptsize


}
\balancecolumns
\end{document}